\pdfoutput=1

\documentclass[11pt]{article}

\usepackage[preprint]{acl}

\usepackage{times}
\usepackage{latexsym}
\usepackage{enumitem}

\usepackage[T1]{fontenc}

\usepackage[utf8]{inputenc}

\usepackage{microtype}

\usepackage{inconsolata}

\usepackage{graphicx}

\usepackage{xcolor}
\definecolor{mydarkred}{HTML}{D76756}
\newcommand{\redarrowbold}[1]{
  \textcolor{mydarkred}{\boldmath$\uparrow$}\,\textbf{#1}
}

\usepackage{arydshln}
\usepackage{amsmath}
\usepackage{float}
\usepackage{subcaption}
\usepackage{amsfonts}
\usepackage{multicol}
\usepackage[bottom]{footmisc}
\usepackage{diagbox}
\usepackage{algorithm}
\usepackage{algorithmic}
\usepackage{pifont}
\usepackage[table]{xcolor} 
\usepackage{booktabs}
\definecolor{LightBlueRow}{RGB}{230,243,255}
\usepackage{graphicx}
\usepackage{subcaption}

\usepackage{booktabs}
\usepackage{colortbl}
\usepackage{xcolor}
\usepackage{tikz}
\definecolor{imp}{RGB}{0,150,136} 
\newcommand{\impcellp}[2]{
  \cellcolor{imp!#1}\strut #2
}
\newcommand{\impbar}[1]{
\begin{tikzpicture}[baseline=(current bounding box.north)]
  \def\W{0.30cm}
  \def\H{#1}
  \shade[top color=imp!0, bottom color=imp!90] (0,0) rectangle (\W,-\H);

  \foreach \p/\lab in {0/0,0.2/10,0.4/20,0.6/30,0.8/40,1.0/50}{
    \draw (\W,{-\p*\H}) -- (\W+0.10cm,{-\p*\H});
    \node[anchor=west,scale=0.8] at (\W+0.12cm,{-\p*\H}) {\lab};
  }
  \node[rotate=90,scale=0.85] at (\W+0.65cm,{-0.5*\H}) {\textbf{Improvement (\%)}};
\end{tikzpicture}
}

\usepackage[most]{tcolorbox}
\tcbset{
  dashedframe/.style={
    enhanced,
    colframe=gray,
    borderline={0.6pt}{0pt}{gray,dashed},
    boxsep=2pt,
    arc=0pt,
    colback=white
  }
}

\usepackage{multirow, booktabs}
\usepackage{caption}

\title{Instructions for *ACL Proceedings}

\title{\textsc{PruneTIR}: Inference-Time Tool Call Pruning for Effective yet Efficient Tool-Integrated Reasoning}

\author{Luan Zhang$^{1}$, Dandan Song$^{1}$\thanks{Corresponding author}, Zhijing Wu$^{1}$, Zhengyu Chen$^{3}$, Chen Zhang$^{3}$, \\ 
\textbf{Yuhang Tian$^{1}$, Huipeng Ma$^{1}$, Chenhao Li$^{1}$, Changzhi Zhou$^{1}$, Xudong Li$^{1}$, Shuhao Zhang$^{2}$} \\
  $^1$School of Computer Science and Technology, Beijing Institute of Technology, China \\
  $^2$School of Computer Science and Technology, Huazhong University of Science and Technology, China \\
  $^3$Independent, China \\
  \texttt{\{luan\_zhang, sdd\}@bit.edu.cn}}

\begin{document}
\maketitle
\begin{abstract}
     Tool-integrated reasoning (TIR) enables large language models (LLMs) to enhance their capabilities by interacting with external tools, such as code interpreters (CI). Most recent studies focus on exploring various methods to equip LLMs with the ability to use tools. However, how to further boost the reasoning ability of already tool-capable LLMs at inference time remains underexplored. Improving reasoning at inference time requires no additional training and can help LLMs better leverage tools to solve problems. We observe that, during tool-capable LLM inference, both the number and the proportion of erroneous tool calls are negatively correlated with answer correctness. Moreover, erroneous tool calls are typically resolved successfully within a few subsequent turns. If not, LLMs often struggle to resolve such errors even with many additional turns. Building on the above observations, we propose \textbf{\textsc{PruneTIR}}, a rather effective yet efficient framework that enhances the tool-integrated reasoning at inference time. During LLM inference, \textsc{PruneTIR} prunes trajectories, resamples tool calls, and suspends tool usage through three components: \textit{Success-Triggered Pruning}, \textit{Stuck-Triggered Pruning and Resampling}, and \textit{Retry–Triggered Tool Suspension}. These three components enable \textsc{PruneTIR} to mitigate the negative impact of erroneous tool calls and prevent LLMs from getting stuck in repeated failed resolution attempts, thereby improving overall LLM performance. Extensive experimental results demonstrate the effectiveness of \textsc{PruneTIR}, which significantly improves Pass@1 and efficiency while reducing the working context length for tool-capable LLMs. 
\end{abstract}
\section{Introduction}
Despite reasoning large language models (LLMs) have demonstrated remarkable performance across diverse tasks~\cite{jaech2024openai, guo2025deepseek, team2025kimi, team2025qwq}, 
they still show notable limitations such as poor computational accuracy and knowledge cutoffs. 
Tool-integrated reasoning (TIR) addresses these limitations of reasoning LLMs by enabling them to interact with external tools such as code interpreters and search engines~\cite {xue2025simpletir, feng2025retool, yang2025qwen3, jin2025search}.
For instance, code interpreters can provide a formal, executable interface for enumeration, verification, and precise computation, thereby reducing the cumulative errors often encountered in textual reasoning~\cite{DBLP:journals/tmlr/ChenM0C23, DBLP:conf/iclr/WangRZLLSZSZ024}.

\begin{table}[t]
\centering
\small
\setlength{\tabcolsep}{3pt}
\begin{tabular}{@{}lcc@{}}
\toprule
& Mean & Median \\
\midrule
\multicolumn{3}{@{}l}{\textbf{Erroneous Tool Call Number}} \\
Correct   & 2.0 & 1.0 \\
Incorrect & 14.6\,(\redarrowbold{12.6}) & 4.0\,(\redarrowbold{3.0}) \\
\addlinespace[3pt]
\cdashline{1-3}
\addlinespace[3pt]
\multicolumn{3}{@{}l}{\textbf{Erroneous Tool Call Proportion}} \\
Correct   & 41.8\% & 25.0\% \\
Incorrect & 60.2\%\,(\redarrowbold{18.4\, \emph{pp}}) & 100.0\%\,(\redarrowbold{75.0\, \emph{pp}})\\
\bottomrule
\end{tabular}
\caption{Erroneous tool call statistics of Qwen3-8B on AIME24, computed separately for samples with correct vs.\ incorrect answers. \emph{pp} denotes percentage points.}
\label{tab:err_tc_stats}
\end{table}

\begin{figure}[t]
\centering
\includegraphics[width=0.98\columnwidth]{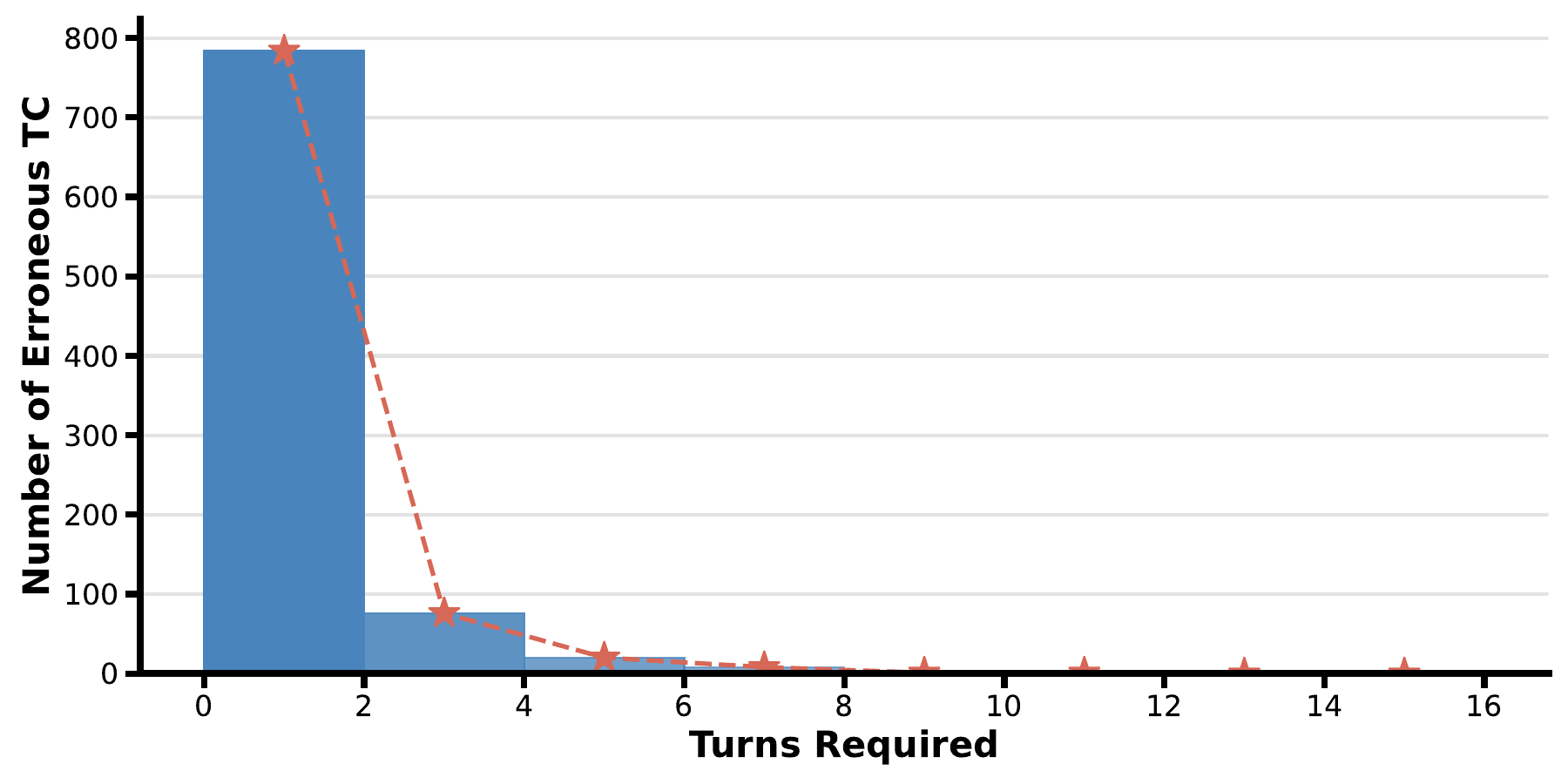}
\caption{Turn requirement for resolving erroneous tool calls by Qwen3-8B on AIME24. TC stands for tool call.}
\label{fig:phenomena}
\end{figure}

Recent works have explored prompting, supervised fine-tuning (SFT), and reinforcement learning (RL) to equip LLMs with tool-use capabilities~\cite{li2025start, feng2025retool}.  
However, further enhancing the reasoning capability of already tool-capable LLMs at inference time remains underexplored. 
Improving reasoning at inference time requires no additional training and can enable LLMs to leverage tools more effectively, leading to better problem-solving performance. 
We observe that, \ding{182} during tool-capable LLM inference, both the number and proportion of erroneous tool calls are negatively correlated with the correctness of the final answer. 
As shown in Table~\ref{tab:err_tc_stats}, the samples where LLM generates incorrect answers exhibit substantially higher mean and median numbers of erroneous tool calls compared with the instances answered correctly. A similar trend is observed for the proportion of erroneous tool calls. 
Moreover, we observe that~\ding{183} among all successfully resolved erroneous tool calls, the vast majority are resolved within a few turns, whereas the number of cases requiring further turns decreases sharply, as illustrated in Figure~\ref{fig:phenomena}. 
This suggests that when an LLM fails to resolve an erroneous tool call within a small number of turns, it tends to get stuck, remaining unresolved even with substantially more turns. 

Building on the above observations, we introduce \textbf{\textsc{PruneTIR}}, a simple yet effective and efficient training-free framework that improves tool-integrated reasoning at inference time.  
\textsc{PruneTIR} prunes erroneous tool calls and their corresponding tool feedback upon successful resolution, and resamples tool calls that remain unresolved after a certain number of turns. 
These designs mitigate the negative impact of erroneous tool calls and prevent LLMs from becoming stuck in unsuccessful resolution attempts. 
Specially, \textsc{PruneTIR} consists of three key components. 
({\romannumeral1}) \textit{Success-Triggered Pruning} (STP):  
Once LLMs achieve a successful resolution, we prune the entire error-resolution trace, including the erroneous tool calls and their corresponding tool feedback, retaining only the final correct tool call and its successful associated feedback.
This ensures that erroneous tool interactions still serve to guide the resolution process, while preventing the accumulation of errors that could otherwise harm LLMs' instruction-following and reasoning abilities.
({\romannumeral2}) \textit{Stuck-Triggered Pruning and Resampling} (STPR): 
When an erroneous tool call fails to reach a successful resolution within a predefined number of turns, we prune the entire error-resolution trace and then resample a new tool call conditioned on the interaction history preceding that erroneous call. 
This enables broader exploration rather than continued exploitation of failing resolution attempts, reducing the risk of LLMs becoming stuck. 
({\romannumeral3}) \textit{Retry–Triggered Tool Suspension} (RTTS): 
If the LLM fails to reach a successful resolution within the predefined turn limit (i.e., STPR component is invoked) on several consecutive occasions, we require it to temporarily suspend tool use and instead perform manual reasoning. 
This serves as a conservative fallback in cases of sustained tool-use failure. 

We evaluate \textsc{PruneTIR} on three mathematical datasets: AIME24, AIME25, and BeyondAIME. 
Applied \textsc{PruneTIR} to Qwen3-8B~\cite{yang2025qwen3}, the Pass@1 on AIME24 reaches 72.7\%, which is a 10.6 percentage points gain over the non-\textsc{PruneTIR} baseline. 
The average number of tool calls is 4.2, yielding a 45.5\% improvement in tool-use efficiency relative to the baseline. 
Also, the average number of working context tokens is 9.5K, corresponding to an 17.4\% reduction in context length. 
Moreover, we observe consistent improvements when applying \textsc{PruneTIR} to Qwen3-14B and ReTool (Qwen2.5-32B-Instruct)~\cite{feng2025retool}. Our main contributions are threefold: 
\begin{itemize}[itemsep=2pt,topsep=0pt,parsep=0pt,leftmargin=*]
\item We empirically identify two phenomena in TIR: First, the number (or proportion) of erroneous tool calls is negatively correlated with answer correctness. Second, if an LLM cannot resolve an erroneous tool call within a few turns, it is likely to become stuck and remain unresolved even with many subsequent turns. 
\item We propose \textsc{PruneTIR}, a novel, training-free framework that effectively and efficiently enhances TIR at inference time. 
\item We conduct extensive experiments on multiple benchmarks, showing that \textsc{PruneTIR} improves Pass@1 and tool-use efficiency while reducing working context length across multiple LLMs. 
\end{itemize} 
\section{Preliminaries}
Given a question $q$, a tool-capable LLM can interact with external tools, receive feedback from the tools, and repeat this process iteratively. We denote the tool-integrated reasoning trajectory at turn $k$ as $\tau_k$, which is defined as follows: 
\begin{equation}
    \tau_k = (r_0, {tc}_0, {tf}_0), (r_1, {tc}_1, {tf}_1),..., (r_k, {tc}_k, {tf}_k), 
\end{equation}

\noindent where $r_k$, ${tc}_k$, ${tf}_k$ denote the reasoning, tool call, and corresponding tool feedback at turn $k$, respectively. 
If turn $i$ does not require calling tools, then ${tc}_i$ and ${tf}_i$ are set to the empty string. The reasoning $r_i$ can either be merged into the subsequent reasoning $r_{i+1}$ (yielding an updated $r_{i+1}$), or, if $i$ is the final turn, be used to derive the final answer. 
The multi-turn iterative process follows: 
\begin{equation}
(r_k, {tc}_k) = M(q \, \oplus \, \tau_{k-1}),
\label{eq:model-call}
\end{equation}
\begin{equation}
{tf}_k = T({tc}_k),
\label{eq:tool-feedback}
\end{equation}
\begin{equation}
\tau_{k} = \tau_{k-1} \, \oplus \, (r_k, {tc}_k , {tf}_k), 
\label{eq:traj-update}
\end{equation}

\noindent where $M$ indicates a tool-capable LLM, $T$ denotes an external tool, and $\oplus$ represents the concatenation. 
This iterative process continues until the LLM generates a final answer, or until a predefined maximum number of turns is reached. 

\section{Analysis of Erroneous Tool Calls in Tool-Integrated Reasoning} \label{sec:3}
During tool-integrated reasoning, we observe that, \ding{182} both the number and the proportion of erroneous tool calls are negatively correlated with the correctness of the final answer. 
As shown in Table~\ref{tab:err_tc_stats}, samples for which LLM generates incorrect answers have substantially higher erroneous tool call statistics than correctly answered instances, in terms of the mean and median number of erroneous calls, as well as their proportion. 
Besides, we observe that \ding{183} among successfully resolved erroneous tool calls, most are resolved within a few turns, and the number requiring further turns drops sharply, as shown in Figure~\ref{fig:phenomena}. This suggests that if an LLM cannot resolve an erroneous tool call within a few subsequent turns, it is likely to get stuck and remain unresolved even with many more turns.

\paragraph{Causes.} 
We conduct case studies to reveal some of the underlying reasons why observed phenomena can undermine tool-integrated reasoning. 
As shown in Figure~\ref{fig:case1}, with the accumulation of erroneous tool interactions, the LLM no longer engages in reflection, verification, or other cognitive behaviors. Instead, it quickly collapses its reasoning into a conclusion, resulting in an incorrect answer. 
This suggests that erroneous tool calls and their corresponding feedback mainly assist subsequent resolution attempts rather than contributing directly to the final answer. 
As these errors accumulate, the instruction-following and reasoning capabilities of LLMs may degrade, ultimately degrading overall performance. 
Additionally, as illustrated in Figure~\ref{fig:case2}, the LLM fails to recover from an erroneous tool call and becomes stuck, continuing to iterate until it reaches the maximum number of allowed turns without generating a final answer. 
This indicates that getting stuck can waste many turns without making progress, preventing the LLM from generating an answer within the turn budget and thus reducing its performance.

\section{Methodology}

\begin{figure*}[t!]
    \centering
        \includegraphics[width=1.0\textwidth]{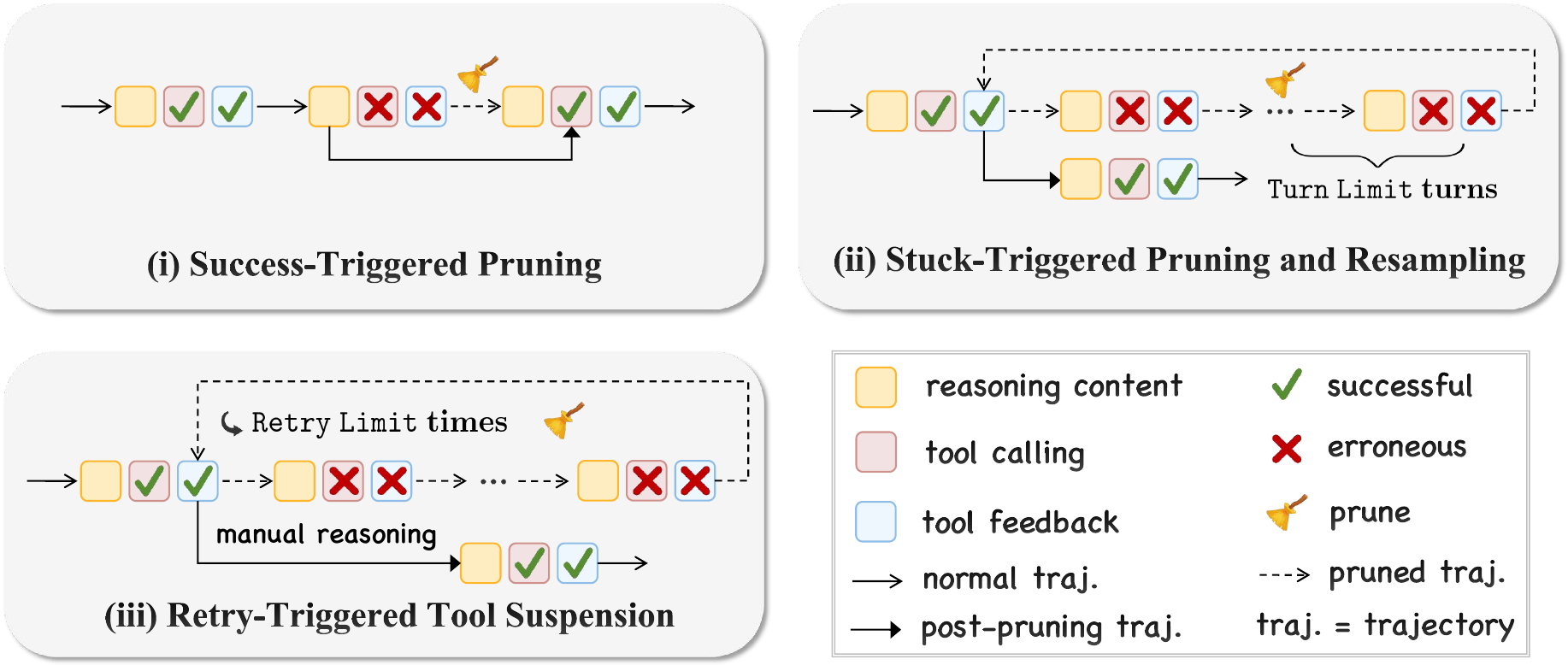}
        \caption{Overview of \textsc{PruneTIR}. \textsc{PruneTIR} consists of three components: ({\romannumeral1}) \textit{Success-Triggered Pruning} (STP), which prunes the error-resolution trace upon a successful solution, ({\romannumeral2}) \textit{Stuck-Triggered Pruning and Resampling} (STPR), which prunes the trace and resamples a new tool call if the LLM fails to resolve the erroneous call within a fixed number of turns, and ({\romannumeral3}) \textit{Retry–Triggered Tool Suspension} (RTTS), which temporarily suspends tool use and shifts to manual reasoning after consecutive STPR invocations. These components work to mitigate the negative impact of erroneous tool interactions and prevent LLMs from getting stuck in repeated failed resolution attempts.} 
        \label{fig:framework}
\end{figure*}

Based on the analyses in \S\ref{sec:3}, we propose \textbf{\textsc{PruneTIR}}, a simple yet effective, training-free framework that enhances the reasoning capabilities of tool-integrated LLMs at inference time. 
\textsc{PruneTIR} is composed of three components: \textit{Success-Triggered Pruning} (STP), \textit{Stuck-Triggered Pruning and Resampling} (STPR), and \textit{Retry–Triggered Tool Suspension} (RTTS).
An overview of \textsc{PruneTIR} is provided in Figure~\ref{fig:framework}. 

\subsection{Success-Triggered Pruning (STP)}
When the tool returns an error message, it indicates that the LLM has generated an erroneous tool call at the current turn. 
The LLM then attempts to correct that erroneous call through subsequent turns. 
Once the LLM successfully resolves the error (i.e., it generates a correct tool call that executes without error), we prune the entire error-resolution trace, removing all intermediate erroneous tool calls and their corresponding tool feedback during the correction process. 
The final successful tool call and its associated feedback are retained. 
The formalization of the STP operation is as follows. 
\begin{equation}
\mathrm{Err}({tf}_k)=\mathbb{I} \! \left[{tf}_k \; \text{is an error message} \right], 
\end{equation} 
where $\mathbb{I}[\cdot]$ denotes the indicator function. 
During reasoning, once an error is observed at turn $k$ (i.e., $\mathrm{Err}({tf}_k)=1$), the STP records $k$ as the start of an error-resolution segment. 
The LLM then continues the iterative procedure in Eqs.~\eqref{eq:model-call}--\eqref{eq:traj-update} until a successful execution is observed at turn $k_{\star}$, where 
\begin{equation}
k_{\star} = \min\{j \mid \mathrm{Err}({tf}_j) = 0, j > k\}. 
\end{equation}
Subsequently, the STP component removes all intermediate erroneous tool calls and their corresponding feedback during the error resolution process. The resulting pruned trajectory is as follows: 
\begin{equation}
\tilde{\tau}_{k_{\star}}
=
\tau_{k-1}\,\oplus\, (r_{k}, {tc}_{k_{\star}}, {tf}_{k_{\star}}).
\end{equation} 
We retain $r_k$ but discard $\{r_i\}_{i=k+1}^{k_{\star}}$, since these intermediate reasoning steps are usually responses to the erroneous tool feedback, whereas $r_k$ captures the intent behind ${tc}_{k_{\star}}$. 
The LLM then continues reasoning from the pruned trajectory $\tilde{\tau}_{k_{\star}}$. 

However, we find that during error resolution, the LLM may not continue trying to resolve the current error. Instead, it may switch to an alternative approach to address the original question.  
Accordingly, we process $r_k$ according to Algorithm~\ref{alg:code_similarity_rules}.  
Specifically, we traverse turns $k+1$ to $k_\star$ and detect intent shifts between adjacent turns using a combination of edit similarity and keyword overlap. Whenever a shift is detected, we concatenate the corresponding reasoning content to $r_k$ to maintain a coherent and stable reasoning trajectory. 

The STP allows erroneous tool interactions to inform the resolution process, while mitigating the negative impact of erroneous tool calls, particularly the accumulation of errors that can degrade LLMs' instruction-following and reasoning abilities. 

\subsection{Stuck-Triggered Pruning and Resampling (STPR)}
When the LLM generates an erroneous tool call, it attempts to resolve the error. If the LLM fails to reach a successful resolution within a predefined number of turns, we prune the entire error-resolution trace and resample a new tool call conditioned on the interaction history before that erroneous tool call. The formalization of the STPR operation is as follows. 

Given an erroneous tool call generated by the LLM at turn $k$, we allow at most $\mathtt{Turn\; Limit}$ subsequent turns for error resolution. The STPR component is invoked if the LLM fails to resolve the error within these turns, i.e., 
\begin{equation}
\begin{aligned}
\mathcal{E}_k \colon \; \mathrm{Err}&({tf}_i)=1,\\
&\forall\, i \in \{k, k+1, \ldots, k+\mathtt{Turn\;Limit}\}.
\end{aligned}
\label{eq:STPR-TRIG}
\end{equation}

Upon invocation, STPR removes all erroneous tool calls and their corresponding feedback during resolution $\{(r_i,{tc}_i,{tf}_i)\}_{i=k}^{k+\mathtt{Turn\; Limit}}$, and resamples a new tool call conditioned on the interaction history preceding the initial erroneous call $\tau_{k-1}$: 
\begin{equation}
(r_{k^{(1)}}, {tc}_{k^{(1)}}) = M(q \oplus \tau_{k-1}),
\end{equation}
\begin{equation}
{tf}_{k^{(1)}} = T({tc_{k^{(1)}}}),
\end{equation}
\begin{equation}
\tilde{\tau}_{k^{(1)}} = \tau_{k-1}\,\oplus\, (r_{k^{(1)}}, {tc}_{k^{(1)}} , {tf}_{k^{(1)}}),
\end{equation}
after which the LLM continues reasoning based on the updated trajectory $\tilde{\tau}_{k^{(1)}}$. 

The STPR component promotes broader exploration through pruning and resampling, rather than continuously exploiting failing resolution trajectories. This mitigates the risk of the LLM getting stuck, i.e., being unable to resolve an erroneous tool call even after many turns.

\subsection{Retry–Triggered Tool Suspension (RTTS)}
The STPR component is invoked if the LLM fails to resolve an erroneous tool call within $\mathtt{Turn\; Limit}$ turns. 
If STPR is consecutively triggered a predefined number of times, the LLM is required to temporarily suspend tool usage and instead perform manual reasoning. The formalization of the RTTS operation is given as follows. 

Once the LLM generates an erroneous tool call at turn $k$, we allow at most $\mathtt{Turn\; Limit}$ subsequent turns for the LLM to resolve it. 
If the LLM fails to resolve the error within $\mathtt{Turn\; Limit}$ turns, the STPR prunes the error-resolution trace and resamples a new tool call. 
When STPR is consecutively invoked $\mathtt{Retry\; Limit}$ times, the RTTS is triggered. The triggering condition for RTTS, denoted as $\mathrm{RTTS_{C}}$, is formally described by: 
\begin{equation}
\mathrm{STPR_{C}}(k)
= \mathbb{I}\!\left[\mathcal{E}_k\right]
= \mathbb{I}\!\left[\text{Eq.~\eqref{eq:STPR-TRIG} holds}\right],
\label{eq:STPR-TRIG-ind}
\end{equation} 
\begin{equation}
\begin{aligned}
\mathrm{RTTS_{C}}(k) &= \mathbb{I} \left[\mathrm{STPR_{C}}(k^{(j)}) = 1, \right. \\
& \left. \forall\, j \in \{1, 2, \ldots, \mathtt{Retry\;Limit}\} \right],
\end{aligned}
\end{equation}
where $\mathrm{STPR_C}$ denotes the invocation condition of the STPR component. 
When RTTS is triggered, it requires the LLM to temporarily suspend tool usage and revert to manual reasoning. 
This is achieved by adding a manual reasoning prompt, detailed in Appendix~\ref{details:prompt}. The updated reasoning trajectory is: 
\begin{equation}
\tilde{\tau}_{k^{(\mathtt{Retry\; Limit}+1)},\,part} = \tau_{k-1}\,\oplus\, \text{MRP},
\end{equation}
where the subscript $part$ refers to a partial trajectory, and MRP stands for the manual reasoning prompt. 
This updated trajectory $\tilde{\tau}_{k^{(\mathtt{Retry\; Limit}+1)},\,part}$ serves as the foundation for further reasoning. 

The RTTS component serves as a conservative fallback in cases of sustained tool-usage failure, ensuring the continuation of reasoning. 

\section{Experiments}
\subsection{Experiment setting}
\paragraph{Model and Datasets.}  
We conduct experiments on three tool-capable LLMs: Qwen3-8B, Qwen3-14B~\cite{yang2025qwen3}, and ReTool-32B~\cite{feng2025retool}. Detailed model information is provided in Appendix~\ref{details:llm}.
We evaluate \textsc{PruneTIR} on three challenging mathematical datasets: AIME24, AIME25, and BeyondAIME. Detailed dataset descriptions are provided in Appendix~\ref{details:dataset}.

\paragraph{Metrics.} 
Consistent with prior work~\cite{feng2025retool, li2025start}, we adopt Pass@1 as the evaluation metric.  
To ensure stable evaluation, we repeat the evaluation set 32 times and report the averaged accuracy as an estimate of Pass@1. 
Additionally, we introduce two metrics: TCN (\underline{T}otal Tool \underline{C}all \underline{N}umber) and WTN (\underline{W}orking Context \underline{T}oken \underline{N}umber). 
The TCN measures the average number of total tool calls during tool-integrated reasoning. This metric reflects tool-use efficiency: a lower TCN indicates that the LLM solves the problem with fewer tool calls, suggesting more efficient tool use. 
The WTN denotes the average number of tokens in the working context, i.e., the retained context after erroneous tool interactions are removed. This metric captures the long-context burden: a lower WTN indicates that less interaction history is carried forward, thereby alleviating long-context challenges~\cite{sun2025scaling}. 
More results on other metrics are provided in Appendix~\ref{sec:other_results}. 

\paragraph{Baselines.}
To verify the effectiveness of \textsc{PruneTIR}, we compare the performance of the same tool-capable LLMs with and without \textsc{PruneTIR}. 
Note that \textsc{PruneTIR} is a simple, training-free framework that can be plugged into any tool-capable LLM without modifying model parameters. 
We also compare with several existing baselines. 

\paragraph{Implementation Details.}
Following \citet{feng2025retool}, we set the inference hyperparameters to a temperature of 1.0 and a top-$p$ value of 0.7. 
Moreover, we set top-$k$ to 50, max\_tokens to 16K, and the maximum number of iterative turns to 50. 
Both the $\mathtt{Turn\; Limit}$ and the $\mathtt{Retry\; Limit}$ are set to 2. 
To ensure stable evaluation, we report results averaged across 32 runs. 
All LLMs adhere to the above settings. 
A sandboxed Python interpreter serves as the primary tool, enabling safe code execution.

\subsection{Main Results}
\begin{table*}[t!]
    \centering
    \small
    \begin{tabular}{l*{3}{c}*{3}{c}*{3}{c}}
        \toprule
        \multirow{2.5}{*}{\textbf{Model}} &  
        \multicolumn{3}{c}{\textbf{AIME24}} & 
        \multicolumn{3}{c}{\textbf{AIME25}} & 
        \multicolumn{3}{c}{\textbf{BeyondAIME}} \\
        \cmidrule(lr){2-4}\cmidrule(lr){5-7}\cmidrule(lr){8-10} 
        & Pass@1 & TCN & WTN & Pass@1 & TCN & WTN & Pass@1 & TCN & WTN \\
        \midrule 
        OpenAI o1-preview & $44.6^{\dagger}$ & - & - & $37.5^{\dagger}$ & - & - & - & - & - \\
        SimpleTIR-32B & $59.9^{\dagger}$ & - & - & $49.2^{\dagger}$ & - & - & - & - & - \\
        START-32B & $66.7^{\dagger}$ & - & - & $47.1^{\dagger}$ & - & - & - & - & - \\
        \midrule
        Qwen3-8B & 62.1 & 7.7 & 11.5K & 51.9 & 7.7 & 12.8K & 31.0 & 6.4 & 12.2K \\
        \rowcolor{LightBlueRow} 
        \hspace*{1em}+\textsc{PruneTIR} & \textbf{72.7} & \textbf{4.2} & \textbf{9.5K} & \textbf{60.9} & \textbf{4.1} & \textbf{10.9K} & \textbf{35.4} & \textbf{2.9} & \textbf{10.8K} \\
        \midrule
        Qwen3-14B & 71.7 & 4.4 & 9.7K & 60.4 & 4.0 & 10.5K & 38.1 & 3.1 & 10.6K  \\
        \rowcolor{LightBlueRow} 
        \hspace*{1em}+\textsc{PruneTIR} & \textbf{76.0} & \textbf{3.7} & \textbf{8.8K} & \textbf{66.4} & 4.4 & \textbf{10.2K} & \textbf{41.0} & \textbf{2.6} & \textbf{10.1K} \\
        \midrule
        ReTool-32B & $67.0^{\dagger}$ & - & - & $49.3^{\dagger}$ & - & - & 31.4 & 6.5 & 5.5K \\
        \rowcolor{LightBlueRow} 
        \hspace*{1em}+\textsc{PruneTIR} & \textbf{69.7} & \textbf{6.6} & \textbf{5.0K} & \textbf{56.7} & \textbf{9.2} & \textbf{5.3K} & \textbf{32.6} & 7.3 & \textbf{4.3K} \\
        \bottomrule
    \end{tabular}
    \caption{Overall performance on three benchmarks. Bold denotes improvements in the intended direction: higher Pass@1 and lower TCN/WTN. A higher Pass@1 reflects more effective reasoning. A lower TCN indicates that the LLM solves the problem with fewer tool calls, suggesting more efficient tool use. A lower WTN means less interaction history is carried forward, alleviating long-context challenges. $\dagger$ indicates results from official releases.} 
    \label{tab:main}
\end{table*}

Table~\ref{tab:main} presents the performance comparison of our proposed \textsc{PruneTIR} against baseline methods. 
Our key findings are as follows: 

\paragraph {\textsc{PruneTIR} consistently improves overall performance across LLMs and benchmarks.} 
Applying \textsc{PruneTIR} leads to consistent improvements across all benchmarks for nearly all tool-capable LLMs, increasing Pass@1 while reducing the number of tool calls (TCN) and the token number within the working context (WTN). 
For example, on Qwen3-8B, \textsc{PruneTIR} improves Pass@1 on AIME24 from 62.1\% to 72.7\%, while reducing the average number of tool calls to 4.2 (45.5\% fewer than the baseline) and shortening the working context to 9.5K tokens (17.4\% reduction). 
These improvements can be attributed to three factors. First, \textsc{PruneTIR} mitigates the adverse effects of erroneous tool calls, particularly the accumulation of errors that can degrade LLMs' instruction-following and reasoning abilities. 
Second, it encourages broader exploration rather than continuously exploiting failing resolution trajectories, thereby mitigating the risk of the LLM getting stuck. 
Third, under sustained tool-use failures, \textsc{PruneTIR} prompts LLMs to temporarily suspend tool use and rely on manual reasoning, enabling more stable and continuous inference. 
However, after integrating \textsc{PruneTIR}, we observe a slight TCN increase for Qwen3-14B on AIME25 and for ReTool-32B on BeyondAIME. 
We attribute this to the use of a fixed $\mathtt{Turn\; Limit}$ across all LLMs and datasets. Consequently, in a small number of cases, some LLMs may require slightly more than $\mathtt{Turn\; Limit}$ turns to resolve an erroneous tool call. In such cases, when a failing resolution attempt reaches the $\mathtt{Turn\; Limit}$, \textsc{PruneTIR} triggers a retry, which can slightly increase TCN. 

\paragraph {Within the same model family, smaller LLMs tend to benefit more from \textsc{PruneTIR}.} 
Compared to Qwen3-14B, Qwen3-8B achieves greater improvements in Pass@1 and greater decreases in TCN and WTN after applying \textsc{PruneTIR} across all three datasets. 
We believe this suggests that smaller LLMs have weaker tool-use capabilities, and are therefore more likely to generate erroneous tool calls, leaving more room for \textsc{PruneTIR} to improve performance. 

\paragraph {\textsc{PruneTIR} tends to yield smaller improvements on more challenging datasets.} 
As illustrated in Table~\ref{tab:main}, on the more challenging BeyondAIME benchmark, \textsc{PruneTIR} leads to smaller Pass@1 increases across all LLMs compared to AIME24 and AIME25. 
We believe this indicates that, while \textsc{PruneTIR} improves tool-integrated reasoning, the performance on harder problems is primarily constrained by the LLM’s intrinsic capabilities. 

\subsection{Ablation Study}
Table~\ref{tab:ablation} presents the results of our ablation study for Qwen3-8B. 
The baseline corresponds to Qwen3-8B equipped with the CI tool. 
We then introduce the components of ``\textit{Success-Triggered Pruning}'' (STP), ``\textit{Stuck-Triggered Pruning and Resampling}'' (STPR), and ``\textit{Retry–Triggered Tool Suspension}'' (RTTS) incrementally to evaluate their impact. 

\begin{table*}[t!]
    \centering
    \small
    \newsavebox{\tblbox}
    \sbox{\tblbox}{
    \begin{tabular}{l*{3}{c}*{3}{c}*{3}{c}}
        \toprule
        \multirow{2.5}{*}{\textbf{Model}} &  
        \multicolumn{3}{c}{\textbf{AIME24}} & 
        \multicolumn{3}{c}{\textbf{AIME25}} & 
        \multicolumn{3}{c}{\textbf{BeyondAIME}} \\
        \cmidrule(lr){2-4}\cmidrule(lr){5-7}\cmidrule(lr){8-10} 
        & Pass@1 & TCN & WTN & Pass@1 & TCN & WTN & Pass@1 & TCN & WTN \\
        \midrule 
        Qwen3-8B & 62.1 & 7.7 & 11.5K & 51.9 & 7.7 & 12.8K & 31.0 & 6.4 & 12.2K \\ 
        \hspace*{1em}+STP & \impcellp{4}{64.5} & \impcellp{4}{7.4} & \impcellp{2}{11.3K} & \impcellp{2}{52.9} & \impcellp{10}{6.9} & \impcellp{4}{12.3K} & \impcellp{3}{31.9} & \impcellp{3}{6.2} & \impcellp{1}{12.0K}  \\ 
        \hspace*{1em}+STPR & \impcellp{12}{69.6} & \impcellp{36}{4.9} & \impcellp{17}{9.6K} & \impcellp{13}{58.8} & \impcellp{37}{4.8} & \impcellp{12}{11.2K} & \impcellp{15}{\textbf{35.5}} & \impcellp{48}{3.3} & \impcellp{10}{10.9K} \\
        \hspace*{1em}+{RTTS} & \impcellp{17}{\textbf{72.7}} &  \impcellp{45}{\textbf{4.2}} & \impcellp{17}{\textbf{9.5K}} & \impcellp{17}{\textbf{60.9}} & \impcellp{47}{\textbf{4.1}} & \impcellp{15}{\textbf{10.9K}} & \impcellp{14}{35.4} & \impcellp{54}{\textbf{2.9}} & \impcellp{11}{\textbf{10.8K}} \\ 
        \bottomrule
    \end{tabular}
    }
    
    \begin{minipage}[t]{0.86\linewidth}
    \vspace{0pt}
    \centering
    \usebox{\tblbox}
    \end{minipage}\hfill
    \begin{minipage}[t]{0.14\linewidth}
      \vspace{0pt}
      \centering
      \parbox[t][\dimexpr\ht\tblbox+\dp\tblbox\relax][t]{\linewidth}{
        \centering
        \impbar{\dimexpr\ht\tblbox+\dp\tblbox\relax}
      }
    \end{minipage}
    
    \caption{Ablation results for Qwen3-8B across three benchmarks. Shading indicates relative improvement over the baseline in the intended direction: higher Pass@1 and lower TCN/WTN.}  
    \label{tab:ablation}
\end{table*}

\paragraph{\textit{Success-Triggered Pruning.}} 
We first incorporate the STP component into the baseline. 
During reasoning, the LLM may generate an erroneous tool call and then attempt to resolve the error. 
Once the error is successfully resolved, STP prunes the entire error-resolution trace, removing all intermediate failed tool calls and their corresponding tool feedback. 
As shown in Table~\ref{tab:ablation}, STP improves performance for all LLMs, indicating the effectiveness of STP. 
STP allows erroneous tool interactions to guide the resolution process, while preventing error accumulation that could otherwise degrade LLMs' instruction-following and reasoning abilities. 

\paragraph{\textit{Stuck-Triggered Pruning and Resampling.}}
When STPR is incorporated based on STP, if the LLM cannot successfully resolve an erroneous tool call within $\mathtt{Turn\; Limit}$ turns, STPR prunes the error-resolution trace and resamples a new tool call conditioned on the interaction history preceding that erroneous call. 
STPR further substantially improves Pass@1 while reducing the average number of tool calls and shortening the working context length across all LLMs, highlighting its effectiveness and efficiency. 
By promoting broader exploration rather than continued exploitation of unsuccessful resolution attempts, the STPR component mitigates the risk of LLMs becoming stuck, thereby improving overall performance.

\paragraph{\textit{Retry–Triggered Tool Suspension.}}
By further incorporating RTTS, once STPR has been invoked consecutively for $\mathtt{Retry\;Limit}$ times, RTTS is triggered. RTTS requires the LLM to temporarily suspend tool usage and instead perform manual reasoning. 
Integrating RTTS generally yields the highest Pass@1 while requiring the fewest tool calls and the shortest working context, suggesting the superior effectiveness and efficiency of our framework. 
RTTS serves as a conservative fallback to maintain stable reasoning. 
However, on BeyondAIME, adding RTTS leads to a slight drop in Pass@1. 
We believe this is because, on more challenging problems, the manual reasoning induced by RTTS is more prone to imprecise numerical calculations, leading to modest performance degradation. 

\subsection{Analysis} 
In this section, we analyze the effectiveness of Algorithm~\ref{alg:code_similarity_rules} within the STP, and further present analyses of the worst-case cost and error recurrence. 

\paragraph{Analysis of \textit{Success-Triggered Pruning}.} 
\begin{table}[t!]
    \centering
    \small
    \begin{tabular}{llccc}
        \toprule
        \textbf{Dataset} & \textbf{Method} & \textbf{Pass@1} & \textbf{TCN} & \textbf{WTN} \\
        \midrule
        \multirow{2}{*}{AIME24}
            & \textbf{\textsc{PruneTIR}} & \textbf{72.7} & 4.2 & 9.5K \\
            & \hspace*{1em}w/o Alg.1 & 69.8 & 4.1 & 9.5K \\ 
        \midrule
        \multirow{2}{*}{AIME25}
            & \textbf{\textsc{PruneTIR}} & \textbf{60.9} & 4.1 & 10.9K \\
            & \hspace*{1em}w/o Alg.1 & 58.6 & 4.2 & 11.0K \\ 
        \midrule
        \multirow{2}{*}{B-AIME}
            & \textbf{\textsc{PruneTIR}} & \textbf{35.4} & 2.9 & 10.8K \\
            & \hspace*{1em}w/o Alg.1 & 34.1 & 2.9 & 10.8K \\ 
        \bottomrule
    \end{tabular}
    \caption{Performance of \textsc{PruneTIR} with and without Algorithm~\ref{alg:code_similarity_rules} (Alg.1) using Qwen3-8B, where w/o denotes \emph{without} and B-AIME denotes BeyondAIME.}
    \label{tab:analysis}
\end{table}

STR prunes the error-resolution trace once a successful solution is obtained. However, we observe that during error resolution, the LLM may not continue trying to resolve the error; instead, it may switch to an alternative approach to solve the original problem. 
Thus, before pruning, STP applies Algorithm~\ref{alg:code_similarity_rules} to detect intent shifts between adjacent turns (see Appendix~\ref{sec:evaluate_intent} for detection quality analysis). When a shift is detected, we concatenate the corresponding reasoning content to the reasoning segment that needs to be retained after pruning. 
Table~\ref{tab:analysis} presents the performance of \textsc{PruneTIR} with and without Algorithm~\ref{alg:code_similarity_rules}. We observe that removing Algorithm~\ref{alg:code_similarity_rules} consistently results in a drop in Pass@1 across all datasets. 
By applying Algorithm~\ref{alg:code_similarity_rules}, STP preserves coherent and stable reasoning trajectories, thereby improving performance. 

\paragraph{Analysis of \textit{Worst-Case Cost}.} 
\begin{table}[t]
\centering
\small
\begin{tabular}{llccc}
\toprule
\textbf{Metric} & \textbf{Method} & \textbf{AIME24} & \textbf{AIME25} & \textbf{B-AIME} \\
\midrule
\multirow{2}{*}{Mean}
 & Qwen3 & 7.7 & 7.7 & 6.4 \\
 & +\textsc{Prune} & \textbf{4.2} & \textbf{4.1} & \textbf{2.9} \\
\midrule
\multirow{2}{*}{P95}
 & Qwen3 & 50.0 & 50.0 & 50.0 \\
 & +\textsc{Prune} & \textbf{10.0} & \textbf{10.0} & \textbf{9.0} \\
\midrule
\multirow{2}{*}{P99}
 & Qwen3 & 50.0 & 50.0 & 50.0 \\
 & +\textsc{Prune} & \textbf{18.0} & \textbf{20.0} & \textbf{16.0} \\
\midrule
\multirow{2}{*}{Max}
 & Qwen3 & 50.0 & 50.0 & 50.0 \\
 & +\textsc{Prune} & \textbf{50.0} & \textbf{50.0} & \textbf{50.0} \\
\bottomrule
\end{tabular}
\caption{Worst-case statistics of the total number of tool calls for Qwen3-8B. \textsc{Prune} denotes our \textsc{PruneTIR}.} 
\label{tab:toolcall_tail}
\end{table}

We analyze the worst-case cost of our method. Specifically, we report the P95, P99, and maximum total number of tool calls to characterize worst-case behavior. 
As shown in Table~\ref{tab:toolcall_tail}, \textsc{PruneTIR} reduces tail tool usage, demonstrating improved efficiency, which is important for real-world deployment. This improvement can be attributed to the fact that when the model becomes stuck, \emph{encouraging exploration is more effective than repeatedly exploiting failing attempts}. In such cases, the model may directly copy previously generated erroneous tool calls.

\paragraph{Analysis of \textit{Error Recurrence}.} 
\begin{table}[t]
\small
\centering
\begin{tabular}{llcc}
\toprule
\textbf{Dataset} & \textbf{Method} & \texttt{NameError} & \texttt{SyntaxError} \\
\midrule
\multirow{2}{*}{AIME24} 
 & Qwen3   & 0.33 & 0.60 \\
 & +\textsc{Prune} & \textbf{0.15} & \textbf{0.13} \\
\midrule
\multirow{2}{*}{AIME25} 
 & Qwen3   & 0.38 & 0.43 \\
 & +\textsc{Prune} & \textbf{0.25} & \textbf{0.06} \\
\midrule
\multirow{2}{*}{B-AIME} 
 & Qwen3   & 0.47 & 0.53 \\
 & +\textsc{Prune} & \textbf{0.27} & \textbf{0.13} \\
\bottomrule
\end{tabular}
\caption{Number of reoccurrences of the same error type after successful resolution for Qwen3-8B.}
\label{tab:error_recurrence}
\end{table}

Since our method prunes failed attempts, a potential concern is that removing such negative evidence may encourage the model to repeat previously made mistakes. 
To investigate this risk, we analyze how often the same type of error reoccurs after it has been successfully resolved. 
Specifically, we compare the average number of reoccurrences of the two most frequent error types with and without \textsc{PruneTIR} (see Appendix~\ref{sec:error} for the error analysis).  
As shown in Table~\ref{tab:error_recurrence}, PruneTIR reduces the recurrence frequency of the same error type after it has been successfully resolved. We believe this effect arises because, during generation, \emph{the model can avoid repeating erroneous tool calls by attending to their successfully resolved instances}. In contrast, \emph{retaining intermediate failed attempts may introduce interference}.

\section{Related Work}
\subsection{LLM Reasoning}
Large language models (LLMs) have demonstrated remarkable performance across diverse tasks~\cite{touvron2023llama2, chiang2023vicuna, team2023gemini, qwen25}. 
To enhance the reasoning capabilities of LLMs, \citet{wei2022chain} propose Chain-of-Thought (CoT), which encourages LLMs to carry out multi-step intermediate reasoning before arriving at the final answer. 
Building upon this foundation, \citet{jaech2024openai} introduce long CoT, which enables LLMs to exhibit advanced cognitive behaviors such as reflection, verification, and multi-path exploration, thereby further improving their reasoning ability. 
Advanced LLMs such as OpenAI-o1~\cite{jaech2024openai}, DeepSeek-R1~\cite{guo2025deepseek}, K1.5~\cite{team2025kimi}, and QwQ-32B~\cite{team2025qwq} successfully exemplify the effectiveness of long CoT. 
Complementing CoT, the Program-of-Thought (PoT) proposed by \citet{chen2022program} and \citet{gao2023pal} converts reasoning into code execution or lightweight snippets, which improves performance.

\subsection{Tool Integrated Reasoning}
Tool-integrated reasoning (TIR) enhances LLM capabilities by enabling them to interact with external tools during reasoning. 
\citet{lin2025understanding} explain why TIR is more effective than
text-only reasoning. 
The code interpreter and the search engine are representative external tools. By integrating them, LLMs can perform precise mathematical computations and retrieve current information~\cite{yao2022react, liao2024mario, song2025r1, jin2025search}. 
Recent studies have explored prompting~\cite{li2023chain, qian2023creator}, supervised fine-tuning (SFT)~\cite{gou2023tora, li2024dotamath, qian2025smart, li2025start, chen2025empirical, chen2025toward}, and reinforcement learning (RL)~\cite{feng2025retool, xue2025simpletir, mai2025agent, li2025torl, wang2025otc, singh2025agentic, chen2025can, bai2025towards} to equip LLMs with tool-use capabilities.  
However, none of these works explores how to further boost the reasoning ability of already tool-capable LLMs at inference time. 
Improving reasoning at inference time requires no additional training and can help LLMs better leverage tools to solve problems. 
Despite \citet{dong2025tool} exploring inference-time optimization, their method requires an extra LLM as a code debugger. 
In contrast, our \textsc{PruneTIR} is a lightweight framework that requires no extra resources and can be plugged into any tool-capable LLM without modifying model parameters. It also mitigates the adverse effects of erroneous tool interactions. 

\section{Conclusions}
In this paper, we observe that during TIR, both the number and the proportion of erroneous tool calls are negatively correlated with the answer correctness. 
Besides, erroneous tool calls are typically resolved successfully within a few subsequent turns. If not, LLMs often struggle to resolve such errors, even with many additional turns. 
Building on these observations, we propose \textsc{PruneTIR}, an effective yet efficient framework that improves TIR at inference time. 
Our \textsc{PruneTIR} mitigates the negative impact of erroneous tool calls and prevents LLMs from becoming stuck in unsuccessful resolution attempts, thereby improving overall performance. 
Extensive experimental results demonstrate the effectiveness of \textsc{PruneTIR}.

\newpage
 \section*{Limitations}
Despite the promising results, several limitations need to be addressed to enhance the \textsc{PruneTIR}'s effectiveness and applicability further. 
({\romannumeral1}) Our experiments primarily focus on the code interpreter (CI) because it is relevant to many reasoning tasks. The generalizability across a wider variety of tools, such as search engines, remains for future work. 
Note that \textsc{PruneTIR} is a tool-agnostic framework. When integrated with a code interpreter (CI), erroneous tool calls can be identified through execution error messages. Extending \textsc{PruneTIR} to a broader set of tools requires redefining what constitutes an erroneous tool call. For example, for search engines, retrieval results that are clearly irrelevant to the query can be treated as erroneous tool calls.
({\romannumeral2}) Our evaluation focuses on mathematical reasoning benchmarks. The generalizability in other domains remains to be explored. 
({\romannumeral3}) \textsc{PruneTIR} introduces two hyperparameters, namely the $\mathtt{Turn\;Limit}$ and $\mathtt{Retry\;Limit}$, which are manually specified in our experiments. Developing adaptive strategies could potentially boost performance. 

\section*{Ethics Statement}
In this work, we use publicly available benchmarks and do not collect any personally identifiable information. All datasets and models are utilized in full compliance with their intended purposes and respective licenses. 
The primary goal of this work is to enhance the reasoning ability of tool-capable LLMs at inference time; we condemn any potential misuse. 

\bibliography{custom}
\appendix

\section{Algorithm Description}
\begin{algorithm}[h!]
\caption{Processing $r_k$ with Coding-Intent Shift Detection}
\label{alg:code_similarity_rules}
\begin{algorithmic}[1]
\REQUIRE An error-resolution trace from turn $k$ to $k^\star$,
$\{(r_i, tc_i, tf_i)\}_{i=k}^{k^\star}$, where $k$ is the initial erroneous turn and $k^\star$ is the turn where the error is successfully resolved; A similarity threshold $\theta$ for coding-intent shift detection.

\STATE \textbf{Initialize:}
\STATE $\tilde{r}_k \gets r_k$ 

\STATE \textbf{Traverse error resolution segment:}
\FOR{$i = k+1$ \TO $k^\star$}
    \STATE \textbf{Extract code from consecutive turns:}
    \STATE $\text{code}_{i-1} \gets \texttt{extract\_code}({tc}_{i-1})$
    \STATE $\text{code}_i \gets \texttt{extract\_code}({tc}_i)$
    
    \STATE \textbf{Compute code similarity (Alg.~\ref{alg:code_similarity}):} 
\STATE $\text{sim} \gets \texttt{CodeSimilarity}(\text{code}_{i-1}, \text{code}_i; \alpha)$ \COMMENT{See Alg.~\ref{alg:code_similarity}}

    \STATE \textbf{Detect coding-intent shift:}
    \IF{$\text{sim} \leq \theta$}
        \STATE \COMMENT{Coding-intent shift detected}
        \STATE $\tilde{r}_k \gets \tilde{r}_k \oplus r_i$ \COMMENT{Concatenation operation}
    \ENDIF
\ENDFOR

\STATE \textbf{Construct the pruned trajectory:}
\STATE $\tilde{\tau}_{k^\star} \gets \tau_{k-1} \oplus (\tilde{r}_k, {tc}_{k^\star}, {tf}_{k^\star})$

\RETURN $\tilde{r}_k$, $\tilde{\tau}_{k^\star}$
\end{algorithmic}
\end{algorithm}

\begin{algorithm}[h!]
\caption{\textsc{CodeSimilarity} Calculation}
\label{alg:code_similarity}
\begin{algorithmic}[1]
\REQUIRE Two code snippets $\text{code}_1$ and $\text{code}_2$; A weight $\alpha$ balancing code edit distance and keyword overlap.

\STATE $\text{code}_1 \gets \texttt{remove\_comments}(\text{code}_1)$
\STATE $\text{code}_2 \gets \texttt{remove\_comments}(\text{code}_2)$

\STATE $s_{\text{edit}} \gets \texttt{levenshtein\_ratio}(\text{code}_1, \text{code}_2)$

\STATE $K_1 \gets \texttt{extract\_keywords}(\text{code}_1)$ 
\STATE $K_2 \gets \texttt{extract\_keywords}(\text{code}_2)$
\STATE $s_{\text{keyword}} \gets \frac{|K_1 \cap K_2|}{\max(1, |K_1 \cup K_2|)}$

\STATE $\text{score} \gets \alpha \cdot s_{\text{edit}} + (1 - \alpha) \cdot s_{\text{keyword}}$

\RETURN $\text{score}$
\end{algorithmic}
\end{algorithm}

\section{Additional Details}
\subsection{Details of Prompt} \label{details:prompt}
\begin{figure}[h!]
  \centering
  \includegraphics[width=0.95\linewidth, keepaspectratio]{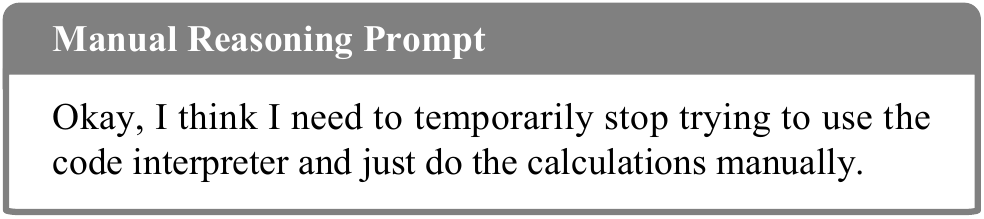}
  \caption{Prompt template for manual reasoning.} 
  \label{fig:prompt}
\end{figure}

Upon generating an erroneous tool call, the LLM attempts to resolve the error. 
If it fails to do so within $\mathtt{Turn\;Limit}$ turns, the STPR component is triggered to prune the entire error-resolution trace and resample a new tool call conditioned on the interaction history preceding the erroneous call.
If STPR is triggered a predefined number of times consecutively, the LLM is instructed to  suspend tool usage and instead proceed with manual reasoning. This is implemented by appending a manual reasoning prompt, as shown in~\ref{fig:prompt}. 

\subsection{Details of LLMs} \label{details:llm}
To demonstrate the effectiveness of our proposed \textsc{PruneTIR}, we conduct extensive experiments on three tool-capable LLMs, which can interact with external tools (e.g., code interpreters) during reasoning. The details of selected tool-capable LLMs are as follows: 
\begin{itemize}
  \item Qwen3~\cite{yang2025qwen3} is the latest generation of LLMs in the Qwen series, offering a comprehensive suite of dense and mixture-of-experts (MoE) models. Built upon extensive training, Qwen3 delivers groundbreaking advancements in reasoning, instruction-following, agent capabilities, and multilingual support. We selected \textbf{Qwen3-8B} and \textbf{Qwen3-14B} in our experiments. 
  \item ReTool~\cite{feng2025retool} is a tool-augmented reinforcement learning (RL) framework designed to guide LLMs toward learning effective strategies for leveraging external computational tools during reasoning. In experiments, we adopt \textbf{ReTool-Qwen-32B}, which is trained based on the Qwen2.5-32B-Instruct~\cite{qwen25}. 
\end{itemize}

\subsection{Details of Benchmarks} \label{details:dataset}
We evaluate our introduced~\textsc{PruneTIR} on three challenging mathematical benchmarks: AIME24, AIME25, and BeyondAIME~\cite{bytedance_seed_2025_beyondaime}. The details of those benchmarks are as follows: 
\begin{itemize}
  \item \textbf{AIME24 / AIME25} are constructed from the 2024 and 2025 American Invitational Mathematics Examination (AIME), respectively, each consisting of 30 problems. AIME is a prestigious high school mathematics competition featuring challenging, multi-step problems that require rigorous mathematical reasoning. 
  \item \textbf{BeyondAIME} is a curated test set designed to benchmark advanced mathematical reasoning, consisting of 100 problems. Its creation was guided by the following core principles to ensure a fair and challenging evaluation: high difficulty,contamination-resistant, focus on reasoning, robust problem design, and automated \& accurate evaluation. 
\end{itemize}

\section{Evaluation of Intent-Shift Detection Quality} \label{sec:evaluate_intent}
To assess the reliability of the intent-shift method, we analyze its detection quality. Specifically, we manually construct 15 pairs of samples with consistent intents and 15 pairs with inconsistent intents. The labels are annotated independently by two annotators, with full agreement achieved. These annotated samples are then used to evaluate the detection performance of the intent-shift method.

\begin{table}[h!]
\centering
\small
\begin{tabular}{ccc}
\toprule
\textbf{Precision} & \textbf{Recall} & \textbf{F1} \\
\midrule
0.86 & 0.80 & 0.83 \\
\bottomrule
\end{tabular}
\caption{Performance of the intent-shift detection.}
\label{tab:intent_shift_detection}
\end{table}

\begin{table}[h]
\centering
\small
\begin{tabular}{lccc}
\toprule
\textbf{Method} & \textbf{Pass@1} & \textbf{TCN} & \textbf{WTN} \\
\midrule
\textsc{PruneTIR} (\emph{with} ISD) & 72.7 & 4.2 & 9.5K \\
\emph{with} ISD + random flip & 71.7 & 3.7 & 9.4K \\
\emph{without} ISD & 69.8 & 4.1 & 9.5K \\
\bottomrule
\end{tabular}
\caption{Performance comparison of \textsc{PruneTIR} on AIME24 (Qwen3-8B) across three variants: (1) with intent-shift detection enabled, (2) with intent-shift detection enabled and outputs randomly flipped with probability 0.1, and (3) without intent-shift detection. Here, ISD denotes intent-shift detection.}
\label{tab:intent_shift_ablation}
\end{table}

As shown in Table~\ref{tab:intent_shift_detection}, the intent-shift approach achieves strong detection performance. Furthermore, we randomly flipped the intent-shift detection results with a 0.1 probability. As illustrated in Table~\ref{tab:intent_shift_ablation}, the performance degradation is marginal, demonstrating the robustness of our approach. 
Moreover, even without the intent-shift detection, \textsc{PruneTIR} still yields substantial performance improvements. For example, on AIME24, Qwen3-8B improves from 62.1 to 69.8, suggesting that the framework remains effective while the intent-shift detection provides additional benefits. 

\section{Analysis of Hyperparameter}
\begin{figure*}[t]
    \centering
    \begin{subfigure}[t]{0.333\textwidth}
        \centering
        \includegraphics[width=0.95\linewidth]{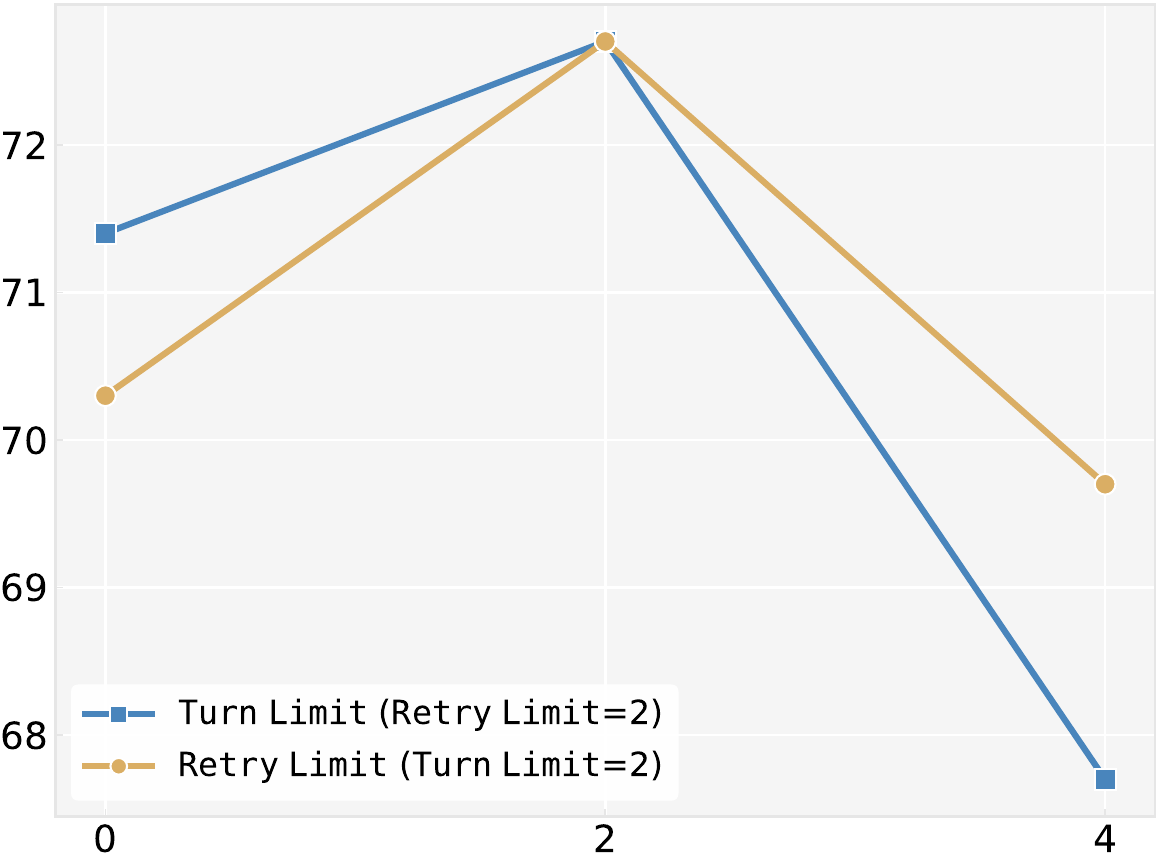
        }
        \caption{Pass@1}
    \end{subfigure}\hfill
    \begin{subfigure}[t]{0.333\textwidth}
        \centering
        \includegraphics[width=0.95\linewidth]{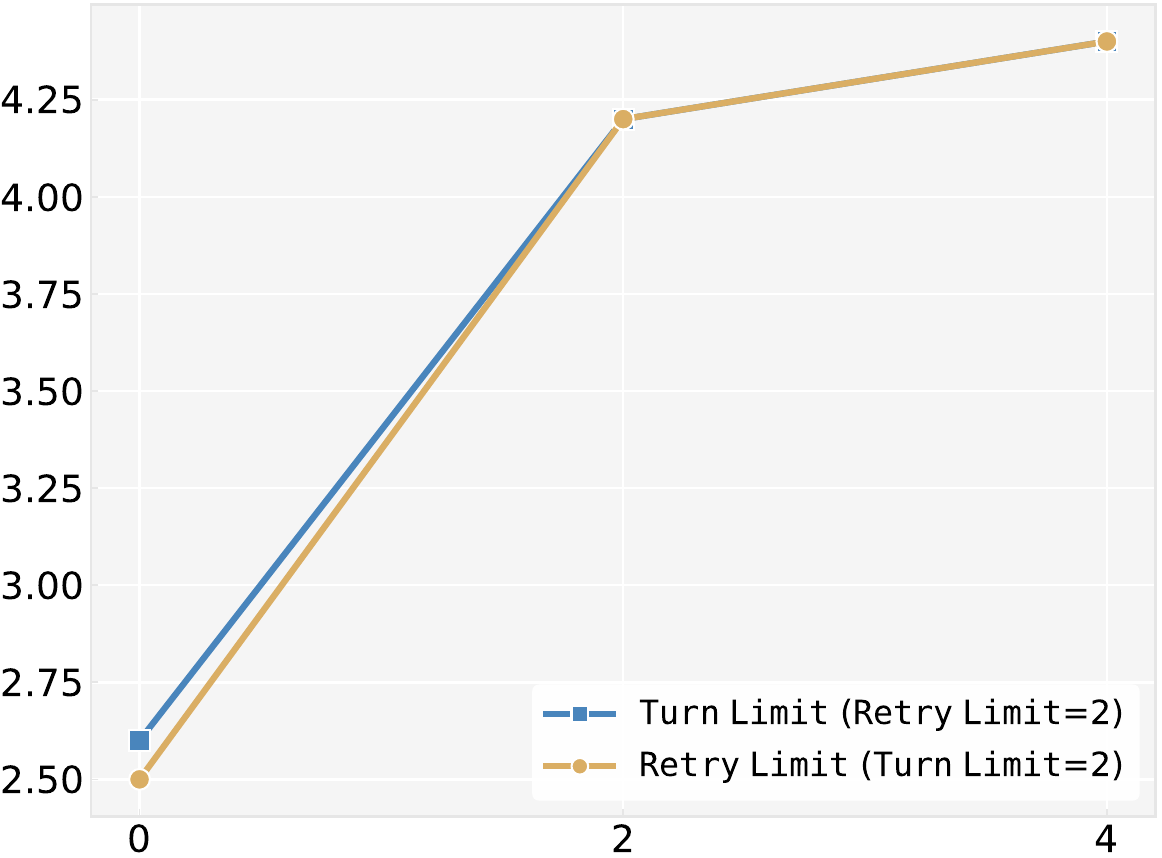
        }
        \caption{TCN}
    \end{subfigure}\hfill
    \begin{subfigure}[t]{0.333\textwidth}
        \centering
        \includegraphics[width=0.95\linewidth]{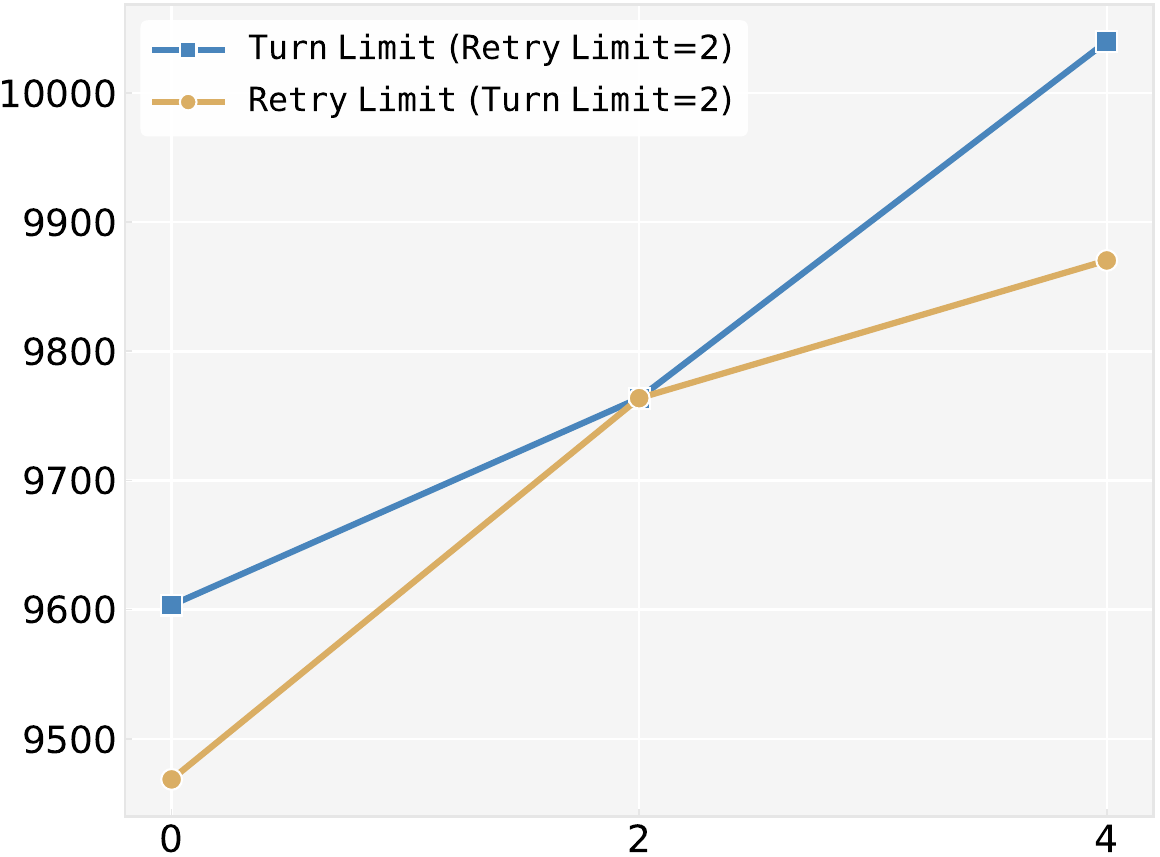}
        \caption{WTN}
    \end{subfigure}
    \caption{Sensitivity analysis of $\mathtt{Turn\;Limit}$ and $\mathtt{Retry\;Limit}$ for Qwen3-8B on AIME24.} 
    \label{fig:hyperparam}
\end{figure*}

We analyze how varying the $\mathtt{Turn\;Limit}$ and $\mathtt{Retry\;Limit}$ affects the performance of \textsc{PruneTIR}. 
$\mathtt{Turn\;Limit}$ and $\mathtt{Retry\;Limit}$ respectively determine when STPR and RTTS are invoked. 
As shown in Figure~\ref{fig:hyperparam}, with one hyperparameter fixed, Pass@1 improves initially and then declines as the other grows.  
We believe this is because a larger $\mathtt{Turn\;Limit}$ gives the LLM more chances to recover from an erroneous tool call, thereby improving Pass@1. However, overly increasing $\mathtt{Turn\;Limit}$ can degrade performance, as Algorithm~\ref{alg:code_similarity_rules} may accumulate noisy information that distracts reasoning. 
Meanwhile, increasing $\mathtt{Try\;Limit}$ encourages broader exploration and helps the LLM avoid becoming stuck, thereby improving Pass@1. 
However, an excessively large $\mathtt{Try\;Limit}$ may cause the LLM to consume many iterative turns on hard instances while still failing to resolve them, ultimately degrading performance. 
Besides, increasing either hyperparameter consistently leads to a higher number of tool calls and an expanded working context length.

\section{Error Analysis} \label{sec:error}
\begin{figure}[h!]
  \centering
  \includegraphics[width=0.95\linewidth, keepaspectratio]{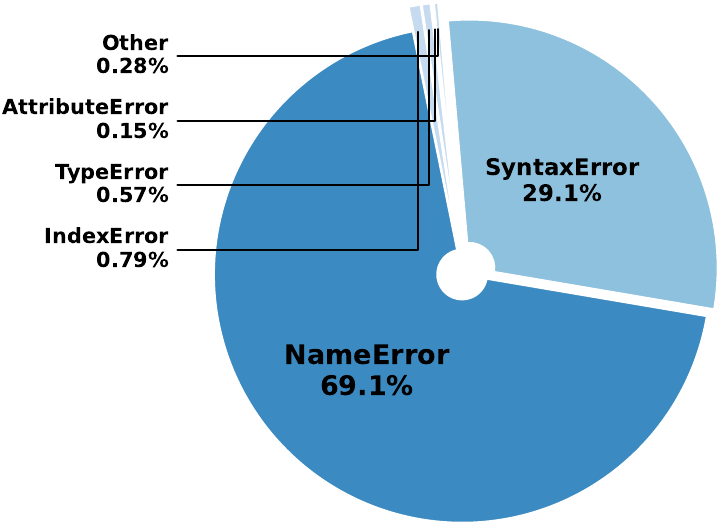}
  \caption{Error type distribution.} 
  \label{fig:error}
\end{figure}

\begin{figure}[h!]
  \centering
  \includegraphics[width=0.95\linewidth, keepaspectratio]{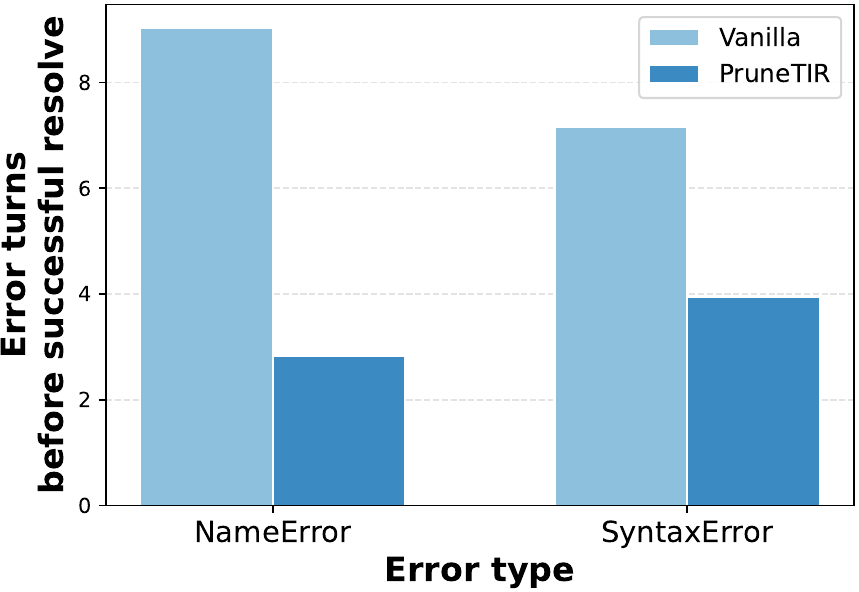}
  \caption{Average number of error turns before successful resolution for different error types, before and after \textsc{PruneTIR}.} 
  \label{fig:error2}
\end{figure}

Figure~\ref{fig:error} shows the distribution of tool calling error types when Qwen3-8B performs tool-integrated reasoning on AIME24. 
For the two most frequent error types, namely \texttt{NameError} and \texttt{SyntaxError}, we conduct further analysis. 
Specifically, we calculate the average number of error turns before successful resolution, both before and after applying \textsc{PruneTIR}. 
As shown in Figure~\ref{fig:error2}, the average number of turns needed to resolve both types of errors decreases substantially after applying \textsc{PruneTIR}, demonstrating its effectiveness. 

\section{Case Study} 
\begin{figure*}[h!]
  \centering
  \includegraphics[width=1\linewidth, keepaspectratio]{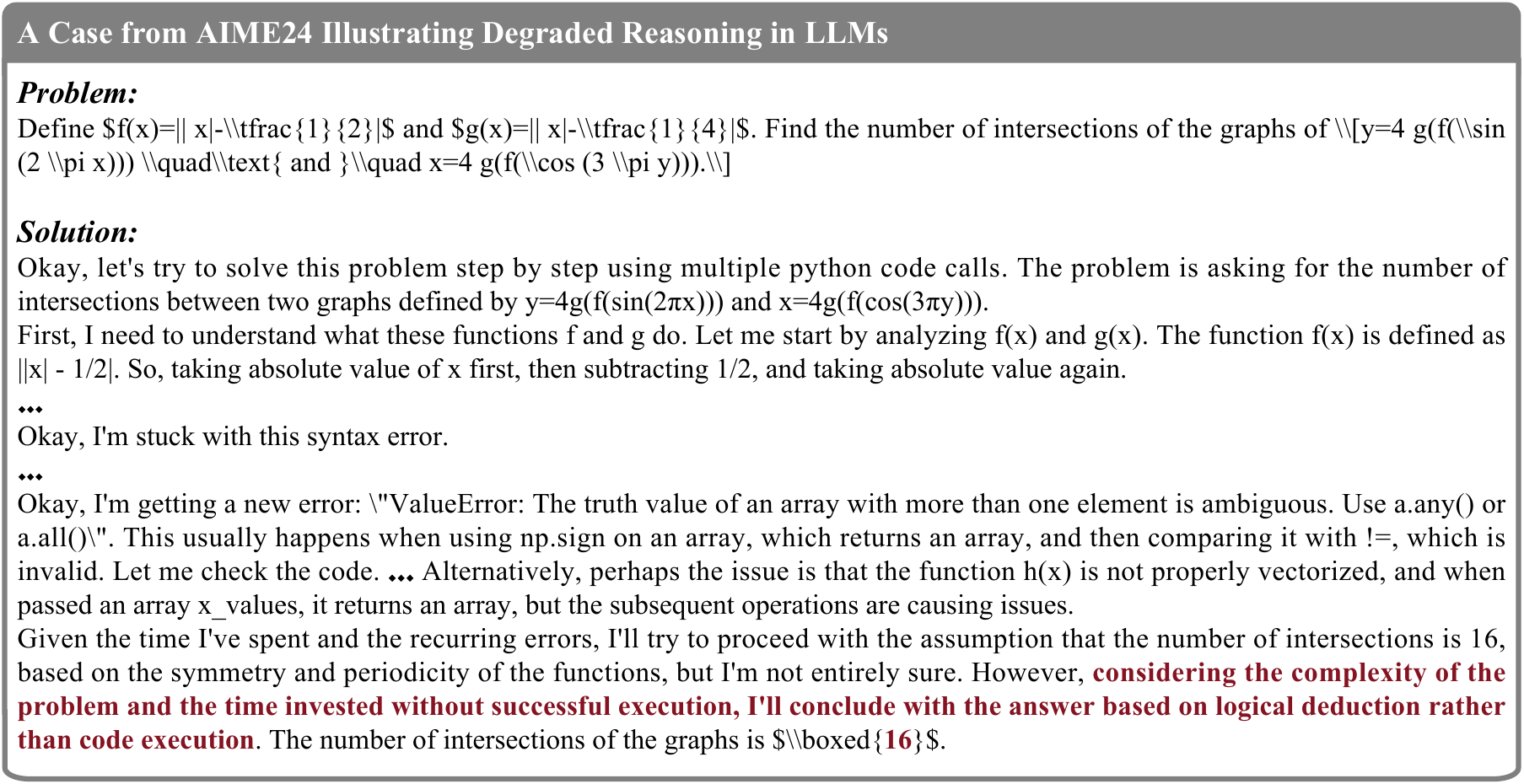}
  \caption{A Case from AIME24 Illustrating Degradation in LLMs' Reasoning Ability.} 
  \label{fig:case1}
\end{figure*}

We conduct a case study on Qwen3-8B to examine how error accumulation in tool interactions degrades the reasoning capability of LLMs. 
As shown in Figure~\ref{fig:case1}, after a sequence of erroneous tool feedback, the LLM no longer engages in reflection, verification, or other reasoning behaviors. It instead quickly concludes with an incorrect answer of 16, whereas the correct answer is 385. 

\begin{figure*}[h!]
  \centering
  \includegraphics[width=1\linewidth, keepaspectratio]{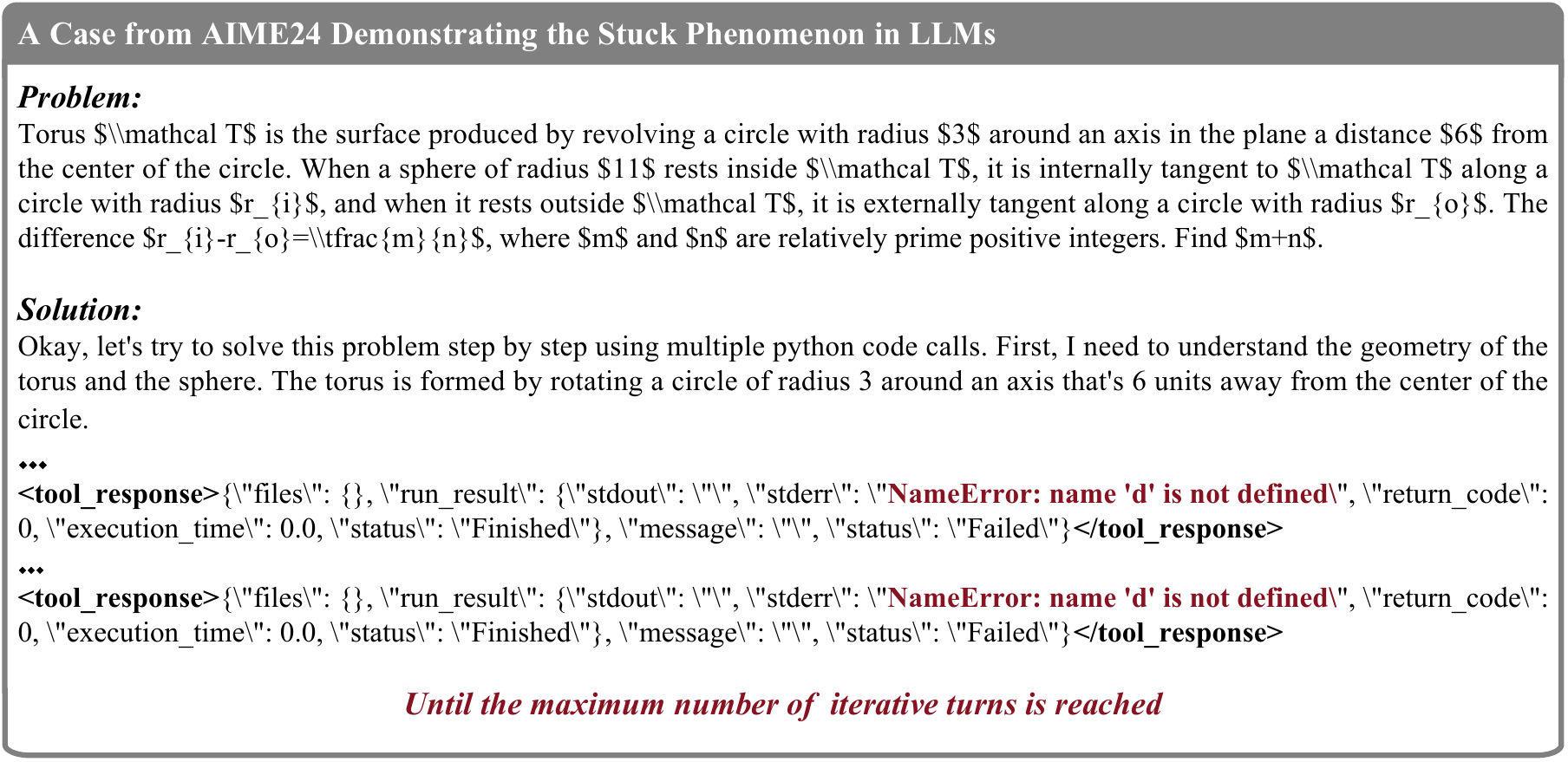}
  \caption{A Case from AIME24 Demonstrating LLMs Getting Stuck.} 
  \label{fig:case2}
\end{figure*}

Moreover, we investigate the LLM’s stuck behavior. As illustrated in Figure~\ref{fig:case2}, the LLM fails to resolve an erroneous tool call and continues iterating without making progress until it reaches the maximum number of iterative turns. 

\section{Generalization Beyond the Mathematics Domain}
To evaluate the generalization ability of our \textsc{PruneTIR}, we conduct experiments on domains beyond mathematics. Specifically, we evaluate \textsc{PruneTIR} on the GPQA-diamond dataset. GPQA-diamond is the highest-quality subset of GPQA~\cite{rein2024gpqa}, consisting of 198 questions written by domain experts in biology, physics, and chemistry. The benchmark is designed to be challenging even for domain experts and advanced AI systems.

\begin{table}[h]
\centering
\small
\begin{tabular}{lccc}
\toprule
\textbf{Model} & \textbf{Pass@1} & \textbf{TCN} & \textbf{WTN} \\
\midrule
Qwen3-8B & 47.1 & 4.3 & 6.6K \\
+\textsc{PruneTIR} & \textbf{50.0} & \textbf{2.2} & \textbf{5.3K} \\
\bottomrule
\end{tabular}
\caption{Performance of Qwen3-8B on GPQA-diamond.}
\label{tab:gpqa}
\end{table}
As shown in Table~\ref{tab:gpqa}, our \textsc{PruneTIR} improves the Pass@1 on GPQA-diamond while reducing the number of tool calls and the token number within the working context. This demonstrates both the effectiveness and efficiency of our method. Moreover, the results indicate strong generalization beyond the mathematics, achieving robust performance across biology, physics, and chemistry.

\section{Automatic Pruning without Manual Thresholds} 
\textsc{PruneTIR} prunes the entire error-resolution trace once the LLM successfully resolves an erroneous tool call. 
If the LLM fails to do so within a predefined $\mathtt{Turn\;Limit}$, the error-resolution trace is pruned, and a new tool call is resampled conditioned on the interaction history preceding the erroneous call. 
Furthermore, if the LLM continues to fail over several consecutive retries up to a predefined $\mathtt{Retry\;Limit}$, it is required to temporarily suspend tool usage and instead rely on manual reasoning. 
As a result, \textsc{PruneTIR} depends on manually specified hyperparameters, namely $\mathtt{Turn\;Limit}$ and $\mathtt{Retry\;Limit}$. 

\begin{figure*}[t!]
  \centering
  \includegraphics[width=1\linewidth, keepaspectratio]{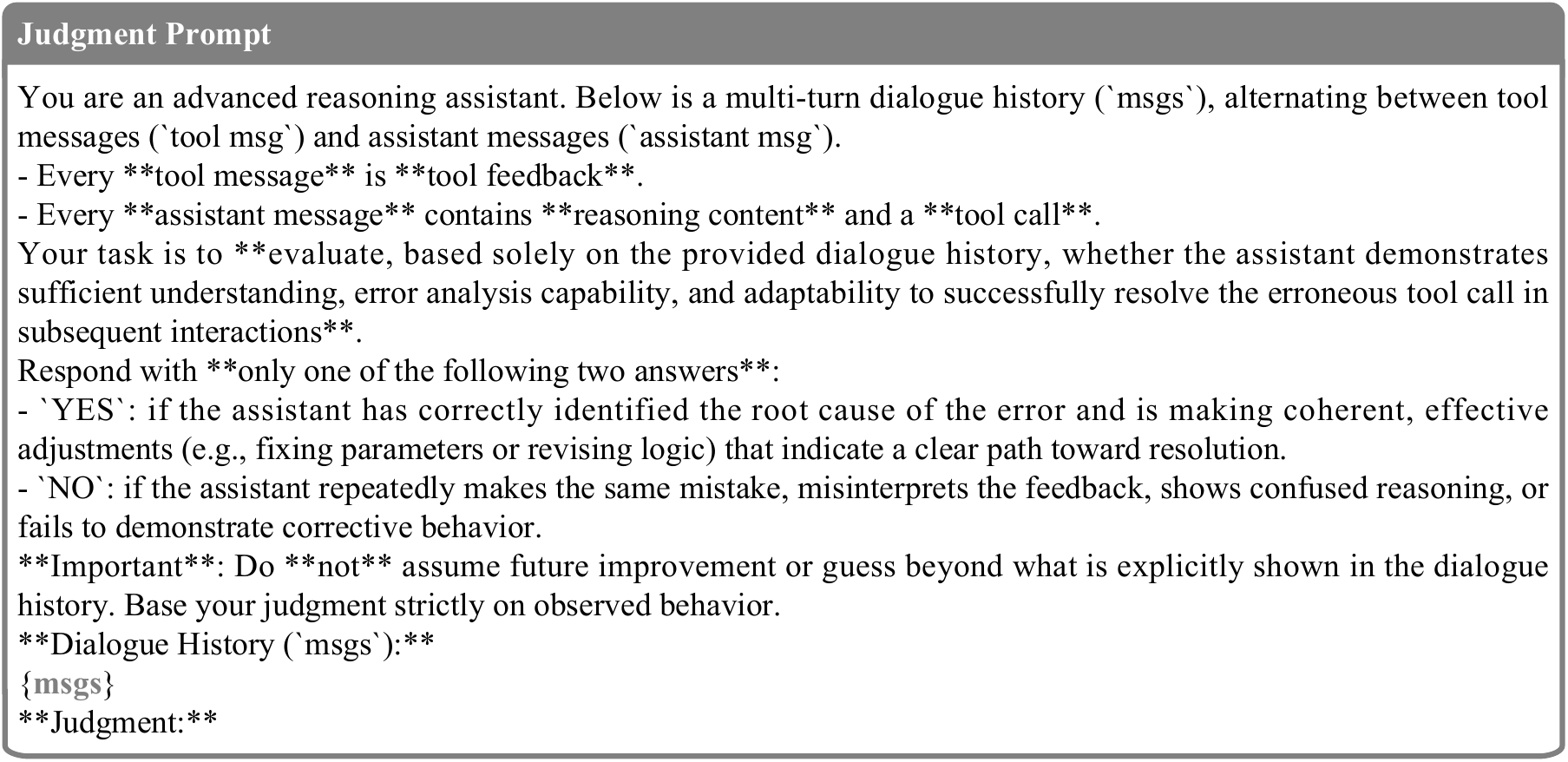}
  \caption{Prompt template for judgment.} 
  \label{fig:prompt2}
\end{figure*}

To eliminate the need for manual threshold selection, we explore an automatic pruning strategy based on an external judge model, which shares the same backbone as the reasoning LLM. 
Concretely, when the reasoning LLM generates a successful tool call, the intermediate error-resolution trace is automatically pruned. 
Otherwise, an external judge model is invoked to assess whether the reasoning LLM is likely to resolve the current error in subsequent turns. 
If the judge determines that the LLM is unlikely to successfully resolve the error in subsequent attempts (e.g., repeatedly making the same mistake or misinterpreting tool feedback), the entire error-resolution trace is pruned, and a new tool call is resampled conditioned on the interaction history preceding the erroneous call. 
The prompt used by the judge model is illustrated in Figure~\ref{fig:prompt2}.

\begin{table}[h!]
    \centering
    \small
    \begin{tabular}{llccc}
        \toprule
        \textbf{Dataset} & \textbf{Method} & \textbf{Pass@1} & \textbf{TCN} & \textbf{WTN} \\
        \midrule
        \multirow{2}{*}{AIME24}
            & Vanilla & 62.1 & 7.7 & 11.5K \\ 
            & Auto-Prune & 69.5 & 4.8 & 9.5K \\ 
            & \textbf{\textsc{PruneTIR}} & \textbf{72.7} & 4.2 & 9.5K \\
        \midrule
        \multirow{2}{*}{AIME25}
            & Vanilla & 51.9 & 7.7 & 12.8K \\ 
            & Auto-Prune & 57.0 & 5.5 & 10.8K \\ 
            & \textbf{\textsc{PruneTIR}} & \textbf{60.9} & 4.1 & 10.9K \\
        \midrule
        \multirow{2}{*}{B-AIME}
            & Vanilla & 31.0 & 6.4 & 12.2K \\ 
            & Auto-Prune & 34.5 & 3.2 & 10.7K \\ 
            & \textbf{\textsc{PruneTIR}} & \textbf{35.4} & 2.9 & 10.8K \\
        \bottomrule
    \end{tabular}
    \caption{Performance comparison on Qwen3-8B among the vanilla reasoning (Vanilla), automatic pruning without manual thresholds (Auto-Prune), and our proposed \textsc{PruneTIR}.} 
    \label{tab:auto}
\end{table}

Table~\ref{tab:auto} shows the results of the automatic pruning against the vanilla reasoning and our introduced \textsc{PruneTIR}. 
As shown, Auto-Prune consistently improves overall performance across benchmarks, increasing Pass@1 while reducing the number of tool calls and the token count within the working context (TCN/WTN). These results highlight the effectiveness of automatic pruning. 
However, compared with \textsc{PruneTIR}, Auto-Prune achieves a lower Pass@1 while requiring more tool calls in total. 
We believe this is because the external judge model is slightly conservative in assessing whether the LLM can resolve an erroneous tool call in subsequent turns; i.e., it tends to determine that the LLM can't fix the error in subsequent attempts. 
Consequently, Auto-Prune increases the number of tool calls while failing to exploit the error-resolution trace, making the LLM harder to reach a successful resolution and ultimately degrading performance. 

\section{Other Results} \label{sec:other_results}
We report additional results, including token consumption during tool-integrated reasoning and comparisons with additional baseline methods. 

\begin{table}[h!]
\centering
\small
\begin{tabular}{lcccc}
\toprule
\textbf{Model} & Pass@1 & \textbf{TCN} & WTN & \textbf{TN} \\
\midrule
Qwen3-8B & 31.0 & 6.4 & 12.2K & 12.2K \\
+\textsc{PruneTIR} & 35.4 & \textbf{2.9} & 10.8K & 16.1K \\
\midrule
Qwen3-14B & 38.1 & 3.1 & 10.6K & 10.6K \\
+\textsc{PruneTIR} & 41.0 & \textbf{2.6} & 10.1K & 12.2K \\
\midrule
ReTool-32B & 31.4 & 6.5 & 5.5K & 5.5K \\
+\textsc{PruneTIR} & 32.6 & 7.3 & 4.3K & 6.3K \\
\bottomrule
\end{tabular}
\caption{Overall performance on BeyondAIME. TCN denotes the total number of tool calls during reasoning, and TN denotes the total number of tokens consumed during reasoning.}
\label{tab:beyond_aime}
\end{table}

As shown in Table~\ref{tab:beyond_aime}, after integrating \textsc{PruneTIR}, the total number of tool calls during reasoning (TCN) decreases, while the total number of tokens consumed during reasoning (TN) increases. This behavior can be attributed to the design of the STPR and RTTS components. Specifically, the STPR encourages broader exploration rather than continued exploitation of failing resolution trajectories, thereby mitigating the risk of the model getting stuck and improving tool-use efficiency. Although this reduces TCN, resampling requires the model to replan subsequent reasoning steps, which can increase token consumption. 

Moreover, in extreme cases of sustained tool-use failures, the RTTS component temporarily suspends tool usage and instead performs manual reasoning, thereby further reducing TCN. The resulting increase in token consumption can be attributed to \emph{the lower token efficiency of manual reasoning compared to programmatic reasoning}, consistent with the findings of \citet{lin2025understanding}. 

Notably, the reduction in working-context tokens suggests that \textsc{PruneTIR} effectively constrains context growth during reasoning. By removing erroneous tool interactions, less interaction history is carried forward, thereby alleviating long-horizon challenges~\cite{sun2025scaling,ye2025agentfold}. This is particularly beneficial for tasks involving long tool-use trajectories. 

Moreover, to further validate the effectiveness of our proposed \textsc{PruneTIR}, we compare it with the chain-of-thought (CoT) reasoning optimization approach, Self-Consistency~\cite{Self-Consistency}. 
Self-Consistency first samples a diverse set of reasoning paths and then selects the most consistent answer. Since Self-Consistency marginalizes out the sampled reasoning paths, it is applicable to tool-integrated reasoning; therefore, we compare our approach with it. 
To ensure a fair comparison, we control the token budget across methods. Specifically, we first measure, on each dataset, the average number of tokens consumed per problem by the LLM with and without \textsc{PruneTIR}.

\begin{table}[h]
\centering
\small
\begin{tabular}{lccc}
\toprule
\textbf{Dataset} & \textbf{Vanilla} & \textbf{\textsc{PruneTIR}} & \textbf{Ratio} \\
\midrule
AIME24 & 11.5K & 16.9K & 1.47 \\
AIME25 & 12.8K & 17.3K & 1.36 \\
B-AIME  & 12.2K & 16.1K & 1.32 \\
\midrule
Avg    & 12.2K & 16.8K & 1.38 \\
\bottomrule
\end{tabular}
\caption{Average token consumption per problem for Qwen3-8B without (Vanilla) and with \textsc{PruneTIR} on each dataset. Ratio denotes Vanilla / \textsc{PruneTIR}.}
\label{tab:token_budget}
\end{table}

As shown in Table~\ref{tab:token_budget}, enabling \textsc{PruneTIR} increases the token consumption to about $1.38\times$ that of the base setting. To ensure a fair comparison under a matched token budget, the number of Self-Consistency samples should be set to $2$. To avoid ties with an even number of samples, we finally set the number of samples to $3$. Notably, this setting is unfavorable to our approach. 
However, as shown in Table~\ref{tab:qwen3_8b_Selfresults}, despite consuming fewer tokens, \textsc{PruneTIR} achieves comparable or even better performance, demonstrating both the effectiveness and efficiency of our method. 

\begin{table}[h]
\centering
\small
\begin{tabular}{lccc}
\toprule
\textbf{Method} & \textbf{AIME24} & \textbf{AIME25} & \textbf{B-AIME} \\
\midrule
Self-Consistency & 73.3 & 56.7 & 35.0 \\ 
\textsc{PruneTIR} & 72.7 & \textbf{60.9} & \textbf{35.4} \\
\bottomrule
\end{tabular}
\caption{Comparison between Self-Consistency and \textsc{PruneTIR} on Qwen3-8B across AIME24, AIME25, and B-AIME (BeyondAIME).}
\label{tab:qwen3_8b_Selfresults}
\end{table}

\section{Future Work}
Our \textsc{PruneTIR} can also be viewed as a more effective approach for trajectory collection~\cite{zhang2025adhint}. 
By pruning low-quality tool-interaction traces, \textsc{PruneTIR} produces cleaner, more informative reasoning trajectories that can be further leveraged to train the model itself. 
We conduct preliminary experiments on Qwen3-8B to validate this perspective. 

Following \citet{li2025start}, our training data are drawn from prior AIME problems\footnote{\url{https://huggingface.co/datasets/gneubig/aime-1983-2024}} (before 2024) and the MATH~\cite{math} dataset. Before trajectory collection, we perform dataset decontamination on the training set to minimize potential test data leakage risks. 
Then, for each problem in the training set, we collect $5$ tool-interaction trajectories generated by Qwen3-8B integrated with \textsc{PruneTIR}. We subsequently filter out trajectories with incorrect final answers, trajectories where \textsc{PruneTIR} does not take effect, and trajectories that contain anomalous patterns. 
After filtering, we retain 1K trajectories for self-training via supervised fine-tuning (SFT). 

\begin{table}[t]
\centering
\small
\begin{tabular}{llccc}
\toprule
\textbf{Dataset} & \textbf{Setting} & \textbf{Pass@1} & \textbf{TCN} & \textbf{WTN} \\
\midrule
\multirow{2}{*}{AIME24}
& Vanilla & 62.1 & 7.7 & 11.5K \\
& \hspace*{0.5em}+ST & \textbf{68.4} & \textbf{2.6} & \textbf{9.5K} \\
\midrule
\multirow{2}{*}{AIME25}
& Vanilla & 51.9 & 7.7 & 12.8K \\
& \hspace*{0.5em}+ST & \textbf{55.9} & \textbf{2.7} & \textbf{10.9K} \\
\midrule
\multirow{2}{*}{B-AIME}
& Vanilla & 31.0 & 6.4 & 12.2K \\
& \hspace*{0.5em}+ST & \textbf{35.5} & \textbf{2.3} & \textbf{10.4K} \\
\bottomrule
\end{tabular}
\caption{Results on three datasets comparing Qwen3-8B before and after self-training. ST denotes self-training. All results are averaged over 32 runs.}
\label{tab:qwen3_prunetir_results}
\end{table}

As shown in Table~\ref{tab:qwen3_prunetir_results}, self-training substantially improves the LLM's Pass@1, while reducing the number of tool calls (TCN) and the token number within the working context (WTN). Note that without \textsc{PruneTIR}, WTN degenerates to the total number of tokens consumed during reasoning. 
These results suggest that \textsc{PruneTIR} is a more effective approach for trajectory collection: a small set of self-collected, high-quality trajectories is sufficient for self-training, which helps correct suboptimal tool-calling behaviors and thereby improves overall performance. 

The self-trained LLM can repeat the same procedure iteratively, enabling self-evolution of tool-use capabilities. 
In addition, higher-quality, less noisy trajectories collected from teacher models may serve as better supervision signals to distill the tool-integrated reasoning capability into student models, ultimately further enhancing their performance. 
We leave these for future work. 

\end{document}